\pdfoutput=1

\documentclass[11pt]{article}

\usepackage[]{naacl2021}

\usepackage{times}
\usepackage{latexsym}

\usepackage[T1]{fontenc}

\usepackage[utf8]{inputenc}

\usepackage{microtype}

\usepackage[linesnumbered, ruled]{algorithm2e}
\usepackage[T1]{fontenc}
\usepackage{dsfont}

\usepackage{mdframed}
\usepackage{booktabs}
\usepackage{graphicx}

\usepackage[normalem]{ulem}
\usepackage{enumitem}
\usepackage{xcolor}
\usepackage{soul}
\colorlet{soulred}{red!30}
\sethlcolor{soulred}%

\usepackage{amsmath,amsthm,amsfonts,amssymb,bm}
\usepackage{framed}
\usepackage{varioref}
\usepackage{float}
\usepackage{multirow}
\usepackage{dashrule}
\usepackage{url}
\usepackage{pifont}
\usepackage[colorinlistoftodos,prependcaption,textsize=tiny]{todonotes}

\newcounter{lh}

\newcounter{xd}

\newcounter{dy}

\newcounter{zy}

\newcounter{ice}
\newcommand{\eat}[1]{\ignorespaces}

\title{Few-shot Intent Classification and Slot Filling with Retrieved Examples}

\author{
Dian Yu\thanks{work done during internship at Google Research}\\
University of California, Davis\\
dianyu@ucdavis.edu
\And
Luheng He \\
Google Research\\
luheng@google.com
\And 
Yuan Zhang\\
Google Research\\
zhangyua@google.com
\AND 
Xinya Du\footnotemark[1]\\
Cornell University\\
xd75@cornell.edu
\And 
Panupong Pasupat\\
Google Research\\
ppasupat@google.com
\And 
Qi Li\\
Google Assistant\\
qilqil@google.com

}

\begin{document}
\maketitle

\begin{abstract}
Few-shot learning arises in important practical scenarios, such as when a natural language understanding system needs to learn new semantic labels for an emerging, resource-scarce domain.
In this paper, we explore retrieval-based methods for intent classification and slot filling tasks in few-shot settings.
Retrieval-based methods make predictions based on labeled examples in the retrieval index that are similar to the input, and thus can adapt to new domains simply by changing the index without having to retrain the model. However, it is non-trivial to apply such methods on tasks with a complex label space like slot filling.
To this end, we propose a span-level retrieval method that learns similar contextualized representations for spans with the same label via a novel batch-softmax objective.
At inference time, we use the labels of the retrieved spans to construct the final structure with the highest aggregated score.
Our method outperforms previous systems in various few-shot settings on the CLINC and SNIPS benchmarks.


\end{abstract}

\section{Introduction}

\begin{figure*}[ht]
\centering
\includegraphics[width=0.9\textwidth]{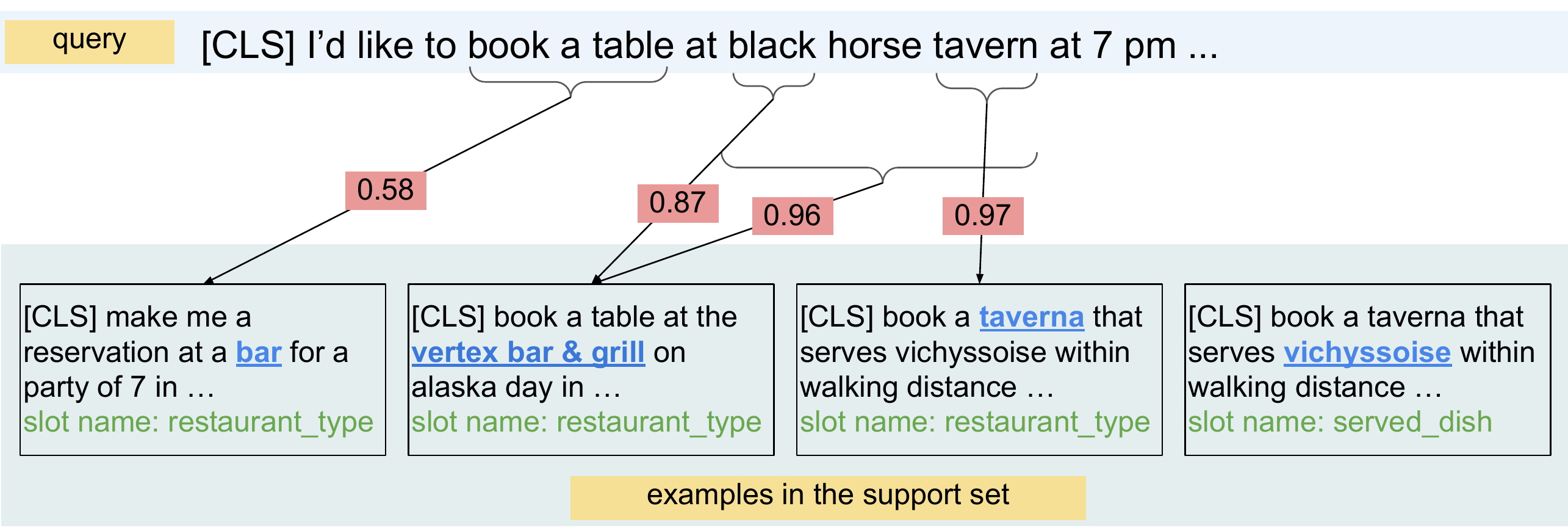}

\caption{Illustration of span-level retrieval for slot filling. 
For each span (including spans that are not valid slots such as \textit{``book a table''}) in the input utterance, we retrieve its most similar span
from the retrieval index,
and then assign the slot name as the prediction with a similarity score. We use modified beam search to decode an output that maximizes the average similarity score. 
The gold slots are \textit{``black horse tavern''} and \textit{``7 pm''} in this example.
}
\label{fig:illustration}
\vspace{-1em}
\end{figure*}

Few-shot learning is a crucial problem for practical language understanding applications.
In the few-shot setting, the model (typically trained on source domains with abundant data) needs to adapt to a set of unseen labels in the target domain with only a few examples. For instance, when developers introduce a new product feature, a query understanding model has to learn new semantic labels from a small dataset they manage to collect.

Few-shot learning is challenging due to the imbalance in the amount of data between the source and target domains.
Traditional classification methods,
even with the recent advancement of
pre-trained language models~\cite{peters-etal-2018-deep, devlin-etal-2019-bert}, could suffer from over-fitting~\cite{snell-etal-2017-prototypical, meta_dataset} or catastrophic forgetting~\cite{wu-etal-2019-transferable}
when incorporating the data-scarce target domain.
On the other hand,
\emph{metric learning methods} \cite{weinberger-etal-2006-distance, vinyals-etal-2016-matching, snell-etal-2017-prototypical} 
have been shown to work well in few-shot scenarios.
These methods are based on modeling similarity between inputs, effectively
allowing the model to be decoupled from the semantics of the output space. For example, a model would learn that the utterance \textit{``I'd like to book a table at black horse tavern at 7 pm''} (from Figure~\ref{fig:illustration}) is similar to \textit{``make me a reservation at 8''} and thus are likely to have similar semantic representations, even without knowing the semantic schema in use.
Unlike learning output labels, which is difficult when examples are scarce,
learning a similarity model can be done on the abundant source domain data, making such models data-efficient even in few-shot settings.

While there are many instantiations of metric learning methods (see Section~\ref{sec:related}), we focus on \emph{retrieval-based} methods, which maintain an explicit retrieval index of labeled examples. The most basic setting of retrieval-based model for few-shot learning is: after training a similarity model and encoding target domain data into the index, we can retrieve examples most similar to the given input, and then make a prediction based on their labels.
Compared to methods that do not maintain an index, such as Prototypical Networks
\cite{snell-etal-2017-prototypical}, retrieval-based methods
are less sensitive to outliers with few data points, and are powerful when we have abundant data in the source domain \cite{meta_dataset}. 

%

However, applying retrieval-based models on tasks with a structured output space is non-trivial.
For example, even if we know that the utterance in Figure~\ref{fig:illustration} is similar to \textit{``make me a reservation at 8''}, we cannot directly use its slot values (e.g., the \texttt{time} slot has value \textit{``8''} which is not in the input), and not all slots in the input (e.g., \textit{``black horse tavern''}) have counterparts in the retrieved utterance.
While previous works have exploited token-level similarity methods in a BIO-tagging framework, they had to separately simulate the label transition probabilities, which might still suffer from domain shift in few-shot settings~\cite{wiseman-stratos-2019-label, hou-etal-2020-shot}.

In this work, we propose \texttt{Retriever}, a retrieval-based framework that tackles both classification and span-level prediction tasks. 
The core idea is to match \emph{token spans} in an input to the most similar labeled spans in a retrieval index.
For example, for the span \textit{``7 pm''} in the input utterance, the model retrieves \textit{``8''} as a similar span (given their surrounding contexts), thus predicting that \textit{``7 pm''} has the same slot name \texttt{time} as \textit{``8''}.
During training, we fine-tune a two-tower model with BERT \cite{devlin-etal-2019-bert} encoders, along with a novel batch softmax objective, to encourage high similarity between contextualized span representations sharing the same label.
At inference time, we retrieve the most similar span from the few-shot examples for every potential input span, and then decode a structured output that has the highest average span similarity score.

We show that our proposed method is effective on both few-shot intent classification and slot-filling tasks, when evaluated on CLINC~\cite{larson-etal-2019-evaluation} and SNIPS~\cite{coucke-etal-2018-snips} datasets, respectively. 
Experimental results show that \texttt{Retriever} achieves high accuracy on few-shot target domains without retraining on the target data. For example, it outperforms the strongest baseline by 4.45\% on SNIPS for the slot-filling task.


\section{Benefits of Retriever}
In addition to being more robust against overfitting and catastrophic forgetting problems, which are essential in few-shot learning settings, our proposed method has multiple advantages overs strong baselines. For instance, if the scheme is changed or some prediction bugs need to be fixed, there is minimum re-training required. More importantly, compared to classification or Prototypical Networks which require adding arbitrary number of instances to the training data and hope that the model will predict as expected \cite{yu-etal-2019-gunrock, liang-etal-2020-gunrock}, \texttt{Retriever} can guarantee the prediction when a similar query is encountered. At the same time, \texttt{Retriever} is more interpretable where the retrieved examples can serve as explanations. In addition, different from the simplified assumption that one utterance may only have one intent \cite{dit++, yu2019midas}, \texttt{Retriever} can be used to predict multiple labels. Lastly, because \texttt{Retriever} does not need to model transition probability, the decoding procedure can be paralleled and potentially be modified to be non-autoregressive for speedup. We can also tune threshold (explained in Section \ref{model_slot_decoding}) to change precision and recall according to use case requirements.

\section{Related Work}
\label{sec:related}

\paragraph{Few-shot metric learning}
Metric learning methods target at learning representations through distance functions. \citet{koch-etal-2015-siamese} proposed Siamese Networks which differentiated input examples with contrastive and triplet loss functions~\cite{Schroffetal-2015-facenet} on positive and negative pairs. 
While they are more data efficient for new classes than linear classifiers, Siamese Networks are hard to train due to weak pairs sampled from training batch \cite{gillick-etal-2019-learning}.
In comparison, Prototypical Networks \cite{snell-etal-2017-prototypical} proposed to compute class representations by averaging embeddings of support examples for each class. 
These methods
have been mostly explored in computer vision and text classification~\cite{geng-etal-2019-induction, yu-etal-2018-diverse}, and 
consistently outperform Siamese Networks and retrieval-based methods such as \textit{k}-nearest-neighbors,  especially when there are more classes and fewer annotated examples \cite{meta_dataset, sun-etal-2019-hierarchical}.
However, newly added examples which are outliers may change the prototypical representations dramatically that can harm all predictions on the class.
In addition, these methods do not perform well when there are more annotated data available per class \cite{meta_dataset}.

Recently, \citet{wang-etal-2019-simpleshot} showed that a simple nearest neighbor model with  feature transformations can achieve competitive results with the state-of-the-art methods on image classification. Inspired by their work, we train our retrieval-based model with a novel batch softmax objective. 

\paragraph{Metric learning in language understanding}
Utilizing relevant examples to boost model performance has been applied to language modeling \cite{khandelwal-etal-2020-generalization}, question answering \cite{guu2020realm, lewis2020retrievalaugmented}, machine translation \cite{zhang-etal-2018-guiding}, and text generation \cite{peng-etal-2019-text}. 
Recently, metric learning has been applied to intent classification \cite{sun-etal-2019-hierarchical, krone-etal-2020-learning}. \citet{ren-etal-2020-intention} trained Siamese Networks before learning a linear layer for intent classification and showed competitive results with traditional methods in the full-data setting. 
Similar ideas are also extended to sequence labeling tasks such as named entity recognition (NER, \citealp{wiseman-stratos-2019-label, fritzler-etal-2018-fewshot}) by maximizing the similarity scores between contextual tokens representations sharing the same label.
\citet{krone-etal-2020-learning} utilized Prototypical Networks to learn intent and slot name prototype representations and classified each token to its closest prototype. They showed better results than meta-learning, another prevalent few-shot learning method \cite{finn-etal-2017-meta, mishra-etal-2018-meta}.
In order to consider label dependencies that are essential in slot tagging tasks \cite{huang-etal-2015-bidirectional}, 
\citet{hou-etal-2020-shot} proposed a collapsed dependency transfer (CDT) mechanism by simulating transition scores for the target domain from transition probabilities among BIO labels in the source domain, outperforming previous methods on slot filling by a large margin. 
\citet{yang-katiyar-2020-simple} further explored the transition probability by evenly distributing the collapsed transition scores to the target domain to maintain a valid distribution. 
However, this simulation is noisy and the difference between the source and target domains can result in biased transition probabilities.

The most similar approach to ours is a concurrent work from~\citet{ziyadi-etal-2020-example}, which learns span boundaries and sentence similarities before retrieving the most similar span, inspired by question-answering models. 
Even though this approach predicts spans before retrieving on the span level and thus bypasses the problem of transition probability in previous research, it only achieves unsatisfactory results. 
Different from these researches, we propose to learn span representations using a batch softmax objective without having to explicitly learn span boundaries. Our method achieves more accurate slot and intent prediction than previous methods in the few-shot setting.

\section{Setup}\label{sec:setup}

We consider two tasks where the input is an utterance $\mathbf{x}$ with tokens $x_1, \dots, x_n$ and the output is some structure $\mathbf{y}$.
For the \emph{slot filling} task, the output $\mathbf{y}$ is a set of non-overlapping labeled spans $\{(r_i, \ell_i)\}_{i=1}^m$ where $r_i$ is a span of $\mathbf{x}$ (e.g., \emph{``7 pm''}) and $\ell_i$ is a slot name (e.g., \texttt{time}). For the \emph{intent classification} task, the output $\mathbf{y}$ is simply an intent label $\ell$ for the whole utterance $\mathbf{x}$. For notational consistency, we view intent classification as predicting a labeled span $(r, \ell)$ where $r = \mathbf{x}_{1:n}$.

In the few-shot setup,
examples $(\mathbf{x}, \mathbf{y})$ are divided into \emph{source} and \emph{target} domains.
Examples in the target domain may contain some labels $\ell$ that are unseen in the source domain.
The model will be given ample training data from the source domain, but only a few training examples from the target domain. For instance, the model receives only $K = 5$ examples for each unseen label.
The model can be evaluated on test data from both domains.

\section{Model}

We propose a retrieval-based model, \texttt{Retriever}, for intent classification and slot filling in the few-shot setting.
Figure \ref{fig:illustration} illustrates our approach.
At a high level, from examples $(\mathbf{x}, \mathbf{y})$ in the target training data (and optionally the source training data), we construct a retrieval index consisting of labeled spans $(r, \ell)$ from $\mathbf{y}$. Given a test utterance $\mathbf{x}$, for each span of interest in $\mathbf{x}$ (all spans $\mathbf{x}_{i:j}$ for slot filling; only $\mathbf{x}_{1:n}$ for intent classification), we retrieve the most similar labeled spans $(r, \ell)$ from the index, and then use them to decode an output $\mathbf{y}$ that maximizes the average span similarity score.

The use of retrieval provides several benefits. For instance, we empirically show in Section \ref{ana:intent} that the model does not suffer from catastrophic forgetting because both source and target data are present in the retrieval index. Class imbalance can also be directly mitigated in the retrieval index. Additionally, since the trained model is non-parametric, we could replace the retrieval index to handle different target domains without having to retrain the model. This also means that the model does not need access to target data during training, unlike traditional classification methods.

\subsection{Retriever}
\label{model:train}

The retriever is the only trainable component in our model. Given a query span $r' = \mathbf{x}_{i:j}$ from the input $\mathbf{x}$, the retriever returns a set of labeled spans $(r, \ell)$ with the highest similarity scores $s(\mathbf{z}, \mathbf{z}')$, where $\mathbf{z} = E(r)$ and $\mathbf{z}' = E(r')$ are the contextualized embedding vectors of $r$ and $r'$, respectively.

\paragraph{Similarity score}
To compute the contextualized embeddings $\mathbf{z}$ and $\mathbf{z}'$ of spans $r$ and $r'$, we first apply a Transformer model initialized with pre-trained BERT on the utterances where $r$ and $r'$ come from. For slot filling, we follow \citet{toshniwal-etal-2020-cross} and define the span embedding as the concatenated embeddings of the its first and last wordpieces.
For intent classification, we use the embedding of the [CLS] token.
We then define $s(\mathbf{z}, \mathbf{z}')$ as the dot product\footnote{We experimented with affine transformation as well as cosine similarity but did not see any performance gain. For intent classification, using the [CLS] token achieves better results than averaging word embeddings.}
between $\mathbf{z}$ and $\mathbf{z}'$.

\paragraph{Training with batch softmax}
We use examples from the source domain to train \texttt{Retriever}. Let $\ell_1, \dots, \ell_N$ be the $N$ class labels (slot or intent labels) in the source domain.
To construct a training batch,
for each class label $\ell_i$, we sample $B$ spans $r_i^1, \dots, r_i^B$ from the training data with that label, and compute their embeddings $\mathbf{z}_i^1, \dots, \mathbf{z}_i^B$.
Then, for each query span $r_i^j$, we compute similarity scores against all other spans in the batch to form a $B\times N$ similarity matrix:
\begin{equation}
\label{eqa:matrix}
S_i^j = \begin{bmatrix}
s(\mathbf{z}_i^j, \mathbf{z}_1^1) & s(\mathbf{z}_i^j, \mathbf{z}_2^1) & \dots & s(\mathbf{z}_i^j, \mathbf{z}_N^1) \\
s(\mathbf{z}_i^j, \mathbf{z}_1^2) & s(\mathbf{z}_i^j, \mathbf{z}_2^2) & \dots & s(\mathbf{z}_i^j, \mathbf{z}_N^2) \\
\vdots & \vdots & \ddots & \vdots \\
s(\mathbf{z}_i^j, \mathbf{z}_1^B) & s(\mathbf{z}_i^j, \mathbf{z}_2^B) & \dots & s(\mathbf{z}_i^j, \mathbf{z}_N^B) \\
\end{bmatrix}.
\end{equation}
We now summarize the score between $r_i^j$ and each label $\ell_{i'}$ by applying a reduction function (defined shortly) along each column to get a $1 \times N$ vector:
\begin{equation}
\label{eqa:prob}
\hat{S}_i^j =
\begin{bmatrix}
s(\mathbf{z}_i^j, \mathbf{z}_1^*) &
s(\mathbf{z}_i^j, \mathbf{z}_2^*) & \dots& s(\mathbf{z}_i^j, \mathbf{z}_N^*)
\end{bmatrix}
\end{equation}
We use the softmax of $\hat{S}_i^j$ as the model's probability distribution on the label of $r_i^j$. The model is then trained to optimize the cross-entropy loss on this distribution against the gold label $\ell_i$.

We experiment with three reduction functions, \textit{mean} (Eq.~\ref{eqa:mean}), \textit{max} (Eq.~\ref{eqa:max}), and \textit{min-max} (Eq.~\ref{eqa:min_max}):

\begin{equation}
\label{eqa:mean}
s(\mathbf{z}_i^j, \mathbf{z}_{i'}^*)
= \frac{1}{B}\sum_{j'=1}^B s(\mathbf{z}_i^j, \mathbf{z}_{i'}^{j'})
= s\Bigg(\mathbf{z}_i^j, \frac{1}{B} \sum\limits_{j'=1}^B \mathbf{z}_{i'}^{j'}\Bigg)
\end{equation}
\begin{equation}
\label{eqa:max}
s(\mathbf{z}_i^j, \mathbf{z}_{i'}^*) = \max_{\substack{1 \leq j' \leq B; \\ j' \neq j \text{ if } i = i'}} s(\mathbf{z}_i^j, \mathbf{z}_{i'}^{j'})
\end{equation}
\begin{equation}
\label{eqa:min_max}
s(\mathbf{z}_i^j, \mathbf{z}_{i'}^*) =
\begin{cases}
    \underset{1 \leq j' \leq B}{\min} s(\mathbf{z}_i^j, \mathbf{z}_{i'}^{j'}),  & \text{if } i = i'\\
    \underset{1 \leq j' \leq B}{\max} s(\mathbf{z}_i^j, \mathbf{z}_{i'}^{j'}),  & \text{otherwise}
\end{cases}
\end{equation}

The \textit{mean} reduction averages embeddings of the spans with the same label and is equivalent to Prototypical Networks. Similar to hard negative sampling to increase margins among classes \cite{Schroffetal-2015-facenet, roth-etal-2020-revisiting, yang-etal-2019-improving}, \textit{max} takes the most similar span to the query (excluding the query itself) as the label representation, while \textit{min-max} takes the least similar span when considering spans with the same label as the query.



\subsection{Inference}\label{sec:inference}


After training, we build a dense retrieval index where each entry $(r, \ell)$ is indexed by $\mathbf{z} = E(r)$.
The entries $(r, \ell)$ come from examples $(\mathbf{x}, \mathbf{y})$ in the \emph{support set} which,
depending on the setting,
could be just the target training data
or a mixture of source and target data.
For each query span $r'$ of the input utterance $\mathbf{x}$, we embed the span and compute the similarity scores against all index entries. 

\paragraph{Intent classification}
For intent classification, both index entries and query spans are restricted to the whole utterances. The entire process thus boils down to retrieving the most similar utterance based on the [CLS] token embedding. We simply output the intent label of the retrieved utterance.

\paragraph{Slot filling}
\label{model_slot_decoding}
In contrast to BIO decoding for token-level similarity models~\cite{hou-etal-2020-shot}, decoding with span retrieval results poses unique challenges as gold span boundaries are not known a priori. Hence, 
we use a modified beam search procedure with simple heuristics to compose the spans.

Specifically, for each of the $n \times m$ spans in an utterance of length $n$ (where the hyperparameter $m$ is the maximum span length), we retrieve the most similar span from the retrieval index.
Then we normalize the similarity scores by L2-norm so that they are within the range $[0, 1]$. 
Since we do not explicitly predict span boundaries, all $n \times m$ spans, including non-meaningful ones (e.g., \textit{``book a''}), will have a retrieved span. Such non-meaningful spans should be dissimilar to any labeled span in the retrieval index. We thus choose to filter the spans with a score threshold to get a smaller set of candidate spans. In addition, we adjust the threshold dynamically (by reducing the threshold for a few times) if no span is above the current threshold.
 
Once we get candidate spans with similarity scores, we use beam search to decode a set of spans with maximum average scores.\footnote{We use beam search for simplicity. Other search methods such as Viterbi algorithm \cite{forney-1973-viterbi} can also be used.}
We go through the list of candidate spans in the descending order of their similarity scores. For each candidate span, we expand beam states if the span does not overlap with the existing spans in the beam.
The search beams are pruned based on the average similarity score of the spans included so far. Lastly, we add spans in the filtered set which do not overlap with the final beam.

Beam search can avoid suboptimal decisions that a greedy algorithm would make.
For instance, if we greedily process the example in Figure~\ref{fig:illustration},
\textit{``black''} and \textit{``tavern''} would become two independent spans, even though their average similarity score is lower than the correct span \textit{``black horse tavern''}.
%
Nevertheless, beam search
is prone to mixing up span boundaries and occasionally predicts consecutive partial spans such as \textit{``black horse''} and \textit{``tavern''} as individual slots.
Since consecutive spans of the \emph{same} slot label are rare in slot filling datasets, we merge the two spans if their retrieval scores are within a certain range:
\begin{equation}
\label{eqa:merge}
\small
    r_{i:j}, r_{j:k} = 
    \begin{cases}
        r_{i:k} & \text{if } |s(\mathbf{z}_{i:j}, \mathbf{z}') - s(\mathbf{z}_{j:k}, \mathbf{z}'')| < \lambda\\
        r_{i:j}, r_{j:k} & \text{otherwise} \nonumber
    \end{cases}
\end{equation}
where $r_{i:j}$ and $r_{j:k}$ are two consecutive potential spans sharing the same label, and $\mathbf{z}'$ and $\mathbf{z}''$ are the embeddings of their retrieved spans, respectively ($r_{i:k}$ indicates merging the two spans into one span; $\lambda$ is the merge threshold where $\lambda = 1$ means always merge and $\lambda = 0$ means never merge).

\begin{table*}[t]
\footnotesize
\begin{center}
\begin{tabular}{l|ccc|ccc|c}
\toprule
& \multicolumn{3}{c|}{support\_set=all} & \multicolumn{3}{c|}{support\_set=balance} & support\_set=tgt \\
\cmidrule(lr){2-4} \cmidrule(lr){5-7} \cmidrule(lr){8-8}
& tgt   & src  & avg   & tgt & src & avg   & tgt  \\ 
\midrule
\multicolumn{4}{l}{\textit{Initial BERT }}     \\
\midrule
$\texttt{Proto}^{\texttt{frz}}$ & 14.07 & 25.02 & 21.37 & - & - & - & -\\
$\texttt{Retriever}^{\texttt{frz}}$ & 8.24 & 54.76 & 39.25 & 22.09 & 25.29 & 24.22 & 37.93\\
\midrule
\multicolumn{4}{l}{\textit{Pre-train on src domain }} \\  
\midrule
\texttt{BERT fine-tune} & - & 96.51 & - & - & - & - & -\\ 
\texttt{Proto} & 75.02 & 95.73 & 88.83 & - & - & - & -\\ 
\texttt{Retriever} & 62.69 & \textbf{97.08} & 85.62 & 75.93 & 95.44 & 88.94 & 88.53\\
\texttt{Retriever  min-max} & 66.00 & 96.64 & 86.43 & 71.82 & 95.14 & 87.37 & 86.38\\
\midrule
\multicolumn{4}{l}{\textit{Fine-tune on tgt domain }} \\
\midrule 
\texttt{BERT fine-tune} & 78.89 & 43.91 & 55.57 & - & - & - & -\\
\texttt{Proto} & 80.44 & 95.57 & 90.53 & - & - & - & 90.35\\
\texttt{Retriever} & 66.76 & 96.95 & 86.89 & 79.20 & 95.50 & 90.07 & \textbf{91.16}\\
\texttt{Retriever  min-max} & 67.64 & 96.84 & 87.11 & 77.60 & 95.35 & 89.43 & 89.56\\
\midrule 
\multicolumn{4}{l}{\textit{Fine-tune on tgt domain with src data}} \\ 
\midrule 
\texttt{BERT  fine-tune} & 72.00 & 95.18 & 87.45 & - & - & - & -\\ 
\texttt{Proto} & 83.33 & 94.82 & 90.99 & - & - & - & 90.22\\
\texttt{Retriever} & 69.51 & 97.04 & 87.86 & \textbf{84.95} & 95.41 & \textbf{91.92} & 90.78\\
\texttt{Retriever  min-max} & 71.35 & 96.96 & 88.42 & 81.00 & 94.55 & 90.03 & 89.82\\

\bottomrule
\end{tabular}
\end{center}
\caption{\label{clinc_results} Intent accuracy on CLINC for $n_c=10, n_i=10$ with 5-shots. Our retrieval-based method outperform BERT fine-tune and Prototypical Networks in both target and source domains. We report results for our method when the support set consists of all examples in the source and target domains (all), when the support set consists of balanced few-shot number of examples for intents in both source and target domains (balance), and when the support set consists of examples of the target domain only (tgt) which serves as an upper-bound.}
\vspace{-1em}
\end{table*}

\section{Experiments and Results}
We evaluate our proposed approach on two datasets: CLINC \cite{larson-etal-2019-evaluation} for intent classification and SNIPS \cite{coucke-etal-2018-snips} for slot filling. Note that we use \textit{max} (Eq.~\ref{eqa:max}) as the reduction function for both tasks since it empirically yields the best results. The effect of reduction functions will be analyzed later in Section~\ref{ana:intent}.

\subsection{Intent Classification}
The CLINC intent classification dataset \cite{larson-etal-2019-evaluation} contains utterances from 10 intent categories (e.g., ``travel''), each containing 15 intents (e.g., ``flight\_status'', ``book\_flight'').
To simulate the few-shot scenario where new domains and intents are introduced, we designate $n_c$ categories and $n_i$ intents per category as the source domain (with all 100 training examples per intent), and use the remaining $150 - n_c \times n_i$ intents as the target domain. We experiment with $(n_c, n_i) = (10, 10)$, $(8, 10)$, and $(5, 15)$.
\footnote{ $(n_c, n_i) = (10, 10)$ simulates the situation where all the categories are known, but we adapt to new intents in all 10 categories; $(n_c, n_i) = (5, 15)$ simulates the situation where we adapt to 5 entirely new intent categories.}
The target training data contains either 2 or 5 examples per target intent.

We compare our proposed method \texttt{Retriever} with
a classification model \texttt{BERT fine-tune} and a Prototypical Network model \texttt{Proto}. The former learns a linear classifier on top of BERT embeddings~\cite{devlin-etal-2019-bert}, and the latter learns class representations based on Prototypical Networks.\footnote{Previous work show that Prototypical Networks outperforms other optimization-based and metric-learning models such as MAML in (intent) classification tasks \cite{meta_dataset, krone-etal-2020-learning}.} We also show results with the initial BERT checkpoint without training ($\texttt{Proto}^{\texttt{frz}}$, $\texttt{Retriever}^{\texttt{frz}}$). We use the same batch size for all models, and tune other hyperparameters on the development set before testing. 

\paragraph{Evaluation}
\label{exp:intent_eval}
We sample domains and intents three times for each $(n_c, n_i)$ setting, and report average prediction accuracy.
We report accuracy on intents from the target domain ({tgt}), source domain ({src}), and the macro average across all intents ({avg}).

In addition to applying the model to the target domain after pre-training on the source domain without re-training (\textit{Pre-train on src domain}), we also evaluate the model performance with fine-tuning. We re-train the model with either target domain data only (\textit{Fine-tune on tgt domain}) or a combination of source and target domain data (\textit{Fine-tune on tgt domain with src data)}.

Moreover, we evaluate the models with the following support set variations: 
with target domain data and all data in the source domain ({support\_set=all}), 
with equal number of examples (same as the few-shot number) per intent ({support\_set=balance}),
and with only examples from the target domain ({support\_set=tgt}). The last one serves as an upper-bound for the target domain accuracy.

\begin{table*}[t]
\footnotesize
\begin{center}
\begin{tabular}{l|ccccccc|c}
\toprule
& GW   & PM    & ATP    & RB    & FSE    & BR    & SCW   & Average F1  \\ 
\midrule
\multicolumn{1}{l}{\textit{Classification-based}}     \\
\midrule
\texttt{BERT Tagging}  & 59.41 & 42.00 & 46.07 & 20.74 & 28.20 & 67.75 & 58.61 & 46.11 \\
\midrule
\multicolumn{1}{l}{\textit{Token-level}}     \\
\midrule
$\texttt{SimilarToken}^{\texttt{frz}}$ & 53.46 & 54.13 & 42.81 & 75.54 & 57.10 & 55.30 & 32.38 & 52.96  \\
\texttt{MatchingToken} & 36.67 & 33.67 & 52.60 & 69.09 & 38.42 & 33.28 & 72.10 & 47.98 \\
\texttt{ProtoToken}  & 67.82 & 55.99 & 46.02 & 72.17 & \textbf{73.59} & 60.18 & \textbf{66.89} & 63.24 \\
\texttt{L-TapNet+CDT+Proto} & - & - & - & - & - & - & - & 67.27\\ 
$\texttt{L-Proto+CDT}^\texttt{pw}$* & 74.68 & 56.73 & 52.20 & 78.79 & 80.61 & 69.59 & 67.46 & 68.58\\
$\texttt{L-TapNet+CDT+Proto}^\texttt{pw}$* & 71.64 & 67.16 & 75.88 & 84.38 & 82.58 & 70.05 & 73.41 & 75.01\\ 

\midrule
\multicolumn{1}{l}{\textit{Span-level (ours)}}     \\
\midrule
$\texttt{Proto}^{\texttt{frz}}$ & 39.47 & 38.35 & 47.68 & 69.36 & 38.60 & 42.39 & 19.90 & 42.25 \\
\texttt{Proto} & 64.47 & 53.97 & 54.64 & 73.37 & 42.89 & 62.48 & 27.76 & 54.23 \\
$\texttt{Retriever}^{\texttt{frz}}$ & 63.39 & 46.01 & 51.11 & 79.65 & 62.42 & 62.13 & 33.85 & 56.94 \\
\texttt{Retriever} & \textbf{82.95} & \textbf{61.74} & \textbf{71.75} & \textbf{81.65} & 73.10 & \textbf{79.54} & 51.35 & \textbf{71.72} \\

\bottomrule
\end{tabular}
\end{center}
\caption{\label{tab:snips_results} Results on SNIPS test data with 5-shot support sets. Our span-based retrieval model outperforms previous classification-based and token-level retrieval models even without label semantics. Classification-based and token-level results are reported in \citet{hou-etal-2020-shot}. *Pair-wise embeddings (marked with $^{\texttt{pw}}$) are expensive at inference time, so we do not compare our method with these directly. 
}
\vspace{-1em}
\end{table*}

\paragraph{Results}
Table \ref{clinc_results} shows the results for $(n_c, n_i) = (10, 10)$ and 5 examples per target intent; results on other settings exhibit the same patterns (See Appendix~\ref{app:clinc_results}). We observe that \texttt{Retriever} performs the best on the source domain (97.08\%) before fine-tuning. \texttt{Retriever} also achieves the highest accuracy on the target domain (84.95\%) after fine-tuning, while maintaining competitive performance on the source domain (95.41\%) among all the methods. 

\subsection{Slot Filling}
SNIPS \cite{coucke-etal-2018-snips} is a slot filling dataset containing 39 slot names from 7 different domains: GetWeather (GW), PlayMusic (PM), AddToPlaylist (ATP), RateBook (RB),  FindScreeningEvent (FSE), BookRestaurant (BR), and SearchCreativeWork (SCW).
Following \citet{hou-etal-2020-shot}, we train models on five source domains, use a sixth one for development, and test on the remaining domain.
We directly use the $K$-shot split provided by \citet{hou-etal-2020-shot}, where the support set consists of the minimum number of utterances such that at least $K$ instances exist for each slot name.
We also set $K = 5$ in our experiment.
Appendix~\ref{app:snips_details} contains further details about the setup.

We compare against two baselines and three models from the previous work. \texttt{BERT Tagging} is a BERT-based BIO tagging model \cite{devlin-etal-2019-bert} fine-tuned on the testing domain after training on the source domains, while $\texttt{SimilarToken}^{\texttt{frz}}$ uses BERT embeddings to retrieve the most similar token based on cosine similarity without any training. 
\texttt{MatchingToken} and \texttt{ProtoToken} are two token-level methods that leveraged Matching Networks \cite{vinyals-etal-2016-matching} and Prototypical Networks \cite{snell-etal-2017-prototypical} 
respectively.
\texttt{L-TapNet+CDT+proto} \cite{hou-etal-2020-shot} is an adaptation of TapNet \cite{yoon-etal-2019-tapent} with label semantics, CDT transition probabilities, and Prototypical Networks.

We experiment with several variants of our proposed method. \texttt{Proto} trains Prototypical Networks to compute span class representations. \texttt{Retriever} retrieves the most similar slot example for each span. Both methods use the same decoding method. Similar to $\texttt{SimilarToken}^{\texttt{frz}}$, $\texttt{Proto}^{\texttt{frz}}$ and $\texttt{Retriever}^{\texttt{frz}}$ use the original BERT embeddings without any training.
%
All models are trained on source domains and early stopped based on performance on the development domains.

\paragraph{Evaluation}
We report F1 scores for each testing domain in a cross-validation episodic fashion. 
Following \citet{hou-etal-2020-shot}, we evaluate each testing domain by sampling 100 different support sets and ten exclusive query utterances for each support set. We calculate F1 scores for each episode and report average F1 scores across 100 episodes.

\paragraph{Results}
Table \ref{tab:snips_results} summarizes the experiment results on the SNIPS dataset. Our span-level method (\texttt{Retriever}) achieves higher averaged F1 than all five baselines, outperforming the strongest token-level method (\texttt{L-TapNet+CDT+proto}) by 4.45\%. This shows that our model is effective at span-level predictions. 
More importantly, the better performance suggests that our span-level \texttt{Retriever} model is more efficient at capturing span structures compared to simulated dependencies as our method does not suffer from the potential discrepancy in the transition probabilities between the target and source domains. 

Although~\citet{hou-etal-2020-shot} showed that adding pairwise embeddings with cross-attention yielded much better performance, this method is expensive both in memory and computation at inference time, especially when the support set is large~\cite{poly_encoder}.
For fair comparison, we do not directly compare with methods using pairwise embeddings (methods with $^\texttt{pw}$ in Table~\ref{tab:snips_results}).
Note that our method with pre-computed support example embeddings even outperforms $\texttt{L-Proto+CDT}^\texttt{pw}$ with less memory and computation cost. 

\section{Analysis}

\subsection{Intent Classification}

\label{ana:intent}

\paragraph{Models without re-training}
The \textit{pre-train on src domain} section in Table \ref{clinc_results} shows the results of models that are only pre-trained on the source domains but \textbf{not} fine-tuned on the target domains. Classification models such as \texttt{BERT fine-tune} cannot make predictions on target domains in this setting. 
In contrast, even without seeing any target domain examples during training, retrieval-based models can still make predictions on new domains by simply including new examples in the support sets. 
With {support\_set=all}, \texttt{Retriever} achieves 97.08\% on the source domain while \texttt{Proto} performs worse than \texttt{BERT fine-tune}, consistent with previous findings \cite{meta_dataset}.
\texttt{Retriever} achieves the best accuracy (75.93\%) on target domains with a balanced support set on all intents ({support\_set=balance}).
More importantly, \texttt{Retriever} also achieves competitive accuracy on source domains (95.44\%), demonstrating that our proposed model achieves the best of both worlds even without re-training on new domains. 

\paragraph{Varying the support set at inference time} 
The construction of the support set is critical to retrieval-based methods.
In Table~\ref{clinc_results}, we present the model performances under different support settings (all, balance, tgt).
The {support\_set=tgt} setting serves as an upper bound for the target domain accuracy for both \texttt{Retriever} and \texttt{Proto} methods.
In general, 
\texttt{Retriever}
achieve the best performance on the source domain intents when we use full support sets (support\_set=all). 
In comparison, if we use a balanced support set (support\_set=balance), we can achieve much higher accuracy on the target domain while having a slight degradation on the source domain intents prediction.
This is because full support sets have more source domain examples to increase confusion over target domains. 
\paragraph{Data for fine-tuning}
The \textit{Fine-tune on tgt domain} section in Table~\ref{clinc_results} shows different model behaviors when fine-tuned on the target domain data directly. 
While \texttt{BERT fine-tune} achieves high accuracy (78.89\%) on the target domain, it suffers from catastrophic forgetting on the source domain (43.91\%).
On the other hand, \texttt{Proto} and \texttt{Retriever} can get high accuracy on the target domain (80.44\% and 79.20\%) while maintaining high performance on the source domain.  

When we combine data from the source domain, we observe performance gains in all the models under the \textit{Fine-tune on tgt domain with src data} section.
Specifically, we add few-shot source domain examples as contrastive examples for the models to learn better utterance/class representations for \texttt{Retriever} and \texttt{Proto}. 
Results show that accuracy on the target domain increases by over 3\% compared to only using target domain data.
This suggests that unlike other retrieval-based methods such as $k$NN, \texttt{Retriever} does not require a large support set to guarantee prediction accuracy.

\paragraph{Impact of reduction functions} 
We compare the reduction functions proposed in Section \ref{model:train} and found that \texttt{max} performs the best.
Since \texttt{mean} is equivalent to Prototypical Networks, we compared to \texttt{Proto} directly in the experiments.
\texttt{min-max} is more intuitive in contrasting with least similar examples within the same class compared to \texttt{max}. However, its performance is worse than \texttt{max}.
We speculate the reason to be that 
we retrieve the example with the maximum score at inference time so that the boundary margin may not be utilized.


\begin{table}[t]
\small
\begin{center}
\resizebox{\columnwidth}{!}{
\begin{tabular}{lccc}
\toprule
& tgt   & src  & avg  \\ 
\midrule
\texttt{BERT fine-tune} & - & - & - \\
\texttt{Proto} & +12.89 & -0.51 & +5.18 \\
\texttt{Retriever} & \textbf{+14.60} & \textbf{-0.14} & \textbf{+6.11}  \\
\texttt{Retriever min-max} & +10.79 & -0.20 & +4.47 \\
\bottomrule
\end{tabular}
}
\end{center}
\caption{\label{clinc_comp_results} Improvement (\%) over \texttt{BERT fine-tune} on target (tgt), source (src), and average (avg), after fine-tuning on the 5-shot support sets. Numbers are averaged over different ($n_c, n_i$) data samples.}
\end{table}

\paragraph{Performance over different settings}
Table~\ref{clinc_comp_results} shows the average improvement of our methods over the \texttt{BERT fine-tune} baseline, where all models are fine-tuned on the target domain with a balanced few-shot dataset after training on the source domain (same as \textit{Fine-tune on tgt domain with src data} section in Table~\ref{clinc_results}). 
Both \texttt{Proto} and \texttt{Retriever} outperforms the baseline on the target domains with a large margin, and \texttt{Retriever} has the best improvement on all intents on average.



\subsection{Slot Filling}
We note that \texttt{Retriever} outperforms the strongest baselines but reaches a low score on the SCW domain. This may be due to the bigger difference between the test (SCW) and the development domain (GW) including the size of the support set and their respective slot names.
We also found that from all the correctly predicted slot spans, 96.73\% predicted the correct slot names. This shows that the majority of the errors come from querying with invalid spans. We believe that span-based pre-training such as Span-BERT \cite{joshi-etal-2020-spanbert} could make our proposed method achieve better results.

\paragraph{Analyzing \texttt{Proto}} 
From Table~\ref{tab:snips_results}, \texttt{Retriever} outperforms \texttt{Proto} by 17\% when training the span representations.
We conjecture that this is caused by \texttt{Proto} learning noisy prototype.
Compared to \texttt{Retriever}, the similarity scores between the spans and their corresponding class representations are low, indicating that the span-level prototypes may not be clearly separated.




\begin{table}[t]
\footnotesize
\begin{center}
\small
\begin{tabular}{ccc}
\toprule
beam size  & merge threshold & avg. F1  \\ 
\midrule
1 & 0.99 & 70.10 \\
5 & 0.99 & 71.47 \\
10 & 0.99 & 71.72 \\
10 & 0 & 70.43 \\
10 & 1 & 72.10 \\
\bottomrule
\end{tabular}
\end{center}
\caption{\label{snips_beam} Ablation study on beam size and merge condition.  Merge threshold of 0 means never merge and 1 means always merge. Using larger beam and merging consecutive spans improve F1 scores.
}
\vspace{-1em}
\end{table}

\paragraph{Ablation on decoding method}
Table \ref{snips_beam} compares beam search to greedy search. Results suggest that beam search with larger beam sizes achieve better F1 scores.
%
As discussed in Section~\ref{sec:inference}, we merge same-label spans during inference based on a score threshold. As shown in Table \ref{snips_beam}, merging spans results in a 1.67\% F1 gain (70.43\% vs 72.10\%) under the same beam size.


\paragraph{Error Analysis}
We find that the main problem of our proposed model is that tokens surrounding the gold span may contain excessive contextual information so that these surrounding invalid spans retrieve corresponding spans with high similarities. For instance, in the query ``add my track to old school metal playlist'', the token ``playlist'' retrieves an actual playlist span with a high similarity score. Another major issue is that the similarity score retrieved by a partial of the gold span sometimes is higher than that retrieved by the whole span. Our ablation results on merge threshold shown in Table \ref{snips_beam} also suggest that partial spans may retrieve complete spans individually so that if we merge consecutive spans with the same slot name, we can achieve higher F1 scores.




\section{Conclusion}
In this paper, we propose a retrieval-based method, \texttt{Retriever}, for few-shot intent classification and slot filling. 
We conduct extensive experiments to compare different model variants and baselines, and show that our proposed approach is effective in the few-shot learning scenario.
We believe that our method can also work on open domain dialog tasks where annotations may be more scarce and other text classification tasks.
In the future, we plan to extend our method to predict more complex structures with span-based retrieval. 

\section{Ethical Considerations}
Our intended use case is few-shot domain adaption to new classes. 
Our experiments are done on English data, but the method is not English-specific.
We use 8 Cloud TPUs V2 cores\footnote{https://cloud.google.com/tpu} for training and one V100 GPU for inference.
Since our model does not have to be retrained for the new domains, it can reduce the resources needed when applying such systems.
We claim that our proposed method outperforms baselines on few-shot slot filling and intent classification examples. Our experiments mainly focus on the 5-shot setting and the 2-shot setting, which are typical testing scenarios applied by previous work with the same claim.
    


\section*{Acknowledgments}
We thank Terry Koo and Emily Pitler from Google Research, and anonymous reviewers for their constructive suggestions.

\bibliography{anthology,custom}
\bibliographystyle{acl_natbib}

\clearpage

\appendix

\section{Appendices}
\label{sec:appendix}

\subsection{Implementation details}
\label{app:implementation}
We use the public uncased BERT-base model from \url{https://github.com/google-research/bert} for embedding spans. Our implementation is adapted from \url{https://github.com/google-research/bert/blob/master/run_classifier.py}. Since the span embedder in the retriever is the only trainable component in our model, the number of parameters is the same as the initial BERT model.

On SNIPS, we set the initial learning rate to be $2 \times 10^{-5}$ with 10\% data for warmup. We set per-class batch size to be 5 for 5-shot experiments.
We use F1 score on the development domain as the metric for early stopping. For decoding, we set $m = 7$ to be the maximum span length and $\lambda=0.99$ as the merging threshold. For dynamic threshold, we decrease the threshold by $0.05$ each time for 10 times until at least one span is above the current threshold.
We also use the development domain results to choose individual threshold for each target domain to filter invalid spans. We use grid search between [0.85, 0.97] with a step of 0.05 to search for the best threshold on the development domain. Our span-level evaluation is modified from \texttt{conlleval} script: \url{https://www.clips.uantwerpen.be/conll2000/chunking/conlleval.txt}.

On CLINC, we set the initial learning rate to be $5 \times 10^{-5}$ and $1 \times 10^{-5}$ for fine-tuning on the target domain. We set per-class batch size to be 8 for training on the source domain, and 5 and 2 for 5-shot and 2-shot fine-tuning.


\subsection{SNIPS Data Details}
\label{app:snips_details}
\begin{table}[h]
\small
\begin{center}
\resizebox{\columnwidth}{!}{
\begin{tabular}{ccc}
\toprule
Test Domain & Dev. Domain & Avg. $|S|$ \\
\midrule
GetWeather & PlayMusic & 28.91 \\
PlayMusic & AddToPlaylist & 34.43 \\
AddToPlaylist & RateBook & 13.84 \\
RateBook & FindScreeningEvent & 19.83 \\
FindScreeningEvent & BookRestaurant & 19.27 \\
BookRestaurant & SearchCreativeWork & 41.58 \\
SearchCreativeWork & GetWeather & 5.28 \\

\bottomrule
\end{tabular}
}
\end{center}
\caption{\label{tab:snips_details} Corresponding development domain and average support set size for each testing domain.}
\end{table}

\subsection{CLINC Results on Different Data Constructions}
\label{app:clinc_results}

\begin{table*}[t]
\footnotesize
\begin{center}
\begin{tabular}{l|ccc|ccc|c}
\toprule
& \multicolumn{3}{c|}{support\_set=all} & \multicolumn{3}{c|}{support\_set=balance} & support\_set=tgt \\
\cmidrule(lr){2-4} \cmidrule(lr){5-7} \cmidrule(lr){8-8}
& tgt   & src  & avg   & tgt & src & avg   & tgt  \\ 
\midrule
\multicolumn{4}{l}{\textit{Initial BERT }}     \\
\midrule
$\texttt{Proto}^{\texttt{frz}}$ & 14.07 & 25.02 & 21.37 & - & - & - & -\\
$\texttt{Retriever}^{\texttt{frz}}$ & 8.24 & 54.76 & 39.25 & 22.09 & 25.29 & 24.22 & 37.93\\
\midrule
\multicolumn{4}{l}{\textit{Pre-train on src domain }} \\  
\midrule
\texttt{BERT fine-tune} & - & 96.51 & - & - & - & - & -\\ 
\texttt{Proto} & 75.02 & 95.73 & 88.83 & - & - & - & -\\ 
\texttt{Retriever} & 62.69 & \textbf{97.08} & 85.62 & 75.93 & 95.44 & 88.94 & 88.53\\
\texttt{Retriever  min-max} & 66.00 & 96.64 & 86.43 & 71.82 & 95.14 & 87.37 & 86.38\\
\midrule
\multicolumn{4}{l}{\textit{Fine-tune on tgt domain }} \\
\midrule 
\texttt{BERT fine-tune} & 78.89 & 43.91 & 55.57 & - & - & - & -\\
\texttt{Proto} & 80.44 & 95.57 & 90.53 & - & - & - & 90.35\\
\texttt{Retriever} & 66.76 & 96.95 & 86.89 & 79.20 & 95.50 & 90.07 & \textbf{91.16}\\
\texttt{Retriever  min-max} & 67.64 & 96.84 & 87.11 & 77.60 & 95.35 & 89.43 & 89.56\\
\midrule 
\multicolumn{4}{l}{\textit{Fine-tune on tgt domain with src data}} \\ 
\midrule 
\texttt{BERT  fine-tune} & 72.00 & 95.18 & 87.45 & - & - & - & -\\ 
\texttt{Proto} & 83.33 & 94.82 & 90.99 & - & - & - & 90.22\\
\texttt{Retriever} & 69.51 & 97.04 & 87.86 & \textbf{84.95} & 95.41 & \textbf{91.92} & 90.78\\
\texttt{Retriever  min-max} & 71.35 & 96.96 & 88.42 & 81.00 & 94.55 & 90.03 & 89.82\\

\bottomrule
\end{tabular}
\end{center}
\caption{\label{clinc_results_10_10_5} Intent accuracy on CLINC for $n_c=10, n_i=10$ with 5-shots.}
\vspace{-1em}
\end{table*}

\begin{table*}[t]
\footnotesize
\begin{center}
\begin{tabular}{l|ccc|ccc|c}
\toprule
& \multicolumn{3}{c|}{support\_set=all} & \multicolumn{3}{c|}{support\_set=balance} & support\_set=tgt \\
\cmidrule(lr){2-4} \cmidrule(lr){5-7} \cmidrule(lr){8-8}
& tgt   & src  & avg   & tgt & src & avg   & tgt  \\ 
\midrule
\multicolumn{4}{l}{\textit{Initial BERT }}     \\
\midrule
$\texttt{Proto}^{\texttt{frz}}$  & 8.70 & 24.06 & 18.94 & - & - & - & -\\
$\texttt{Retriever}^{\texttt{frz}}$ & 3.91 & 54.96 & 37.94 & 17.38 &	16.83 & 17.01 & 27.96\\
\midrule
\multicolumn{4}{l}{\textit{Pre-train on src domain }} \\  
\midrule
\texttt{BERT fine-tune} & - & 96.51 & - & - & - & - & -\\ 
\texttt{Proto} & 76.40 & 95.70 & 89.27 & - & - & - & -\\ 
\texttt{Retriever} & 53.73 & 97.02 & 82.59 & 73.13 & 94.32 & 87.26 & 86.47\\
\texttt{Retriever  min-max} & 53.47 & 96.87 & 82.40 & 68.47 & 95.29 & 86.35 & 81.76\\
\midrule
\multicolumn{4}{l}{\textit{Fine-tune on tgt domain }} \\
\midrule 
\texttt{BERT fine-tune} & 75.57 & 50.91 & 59.13 & - & - & - & -\\
\texttt{Proto} & 76.36 & 95.06 & \textbf{88.82} & - & - & - & 86.67\\
\texttt{Retriever} & 55.20 & 97.05 & 83.10 & 74.89 & 94.57 & 88.01 & 87.74\\
\texttt{Retriever  min-max} & 55.96 & 96.92 & 83.27 & 71.09 & 95.25 & 87.19 & 83.91\\
\midrule 
\multicolumn{4}{l}{\textit{Fine-tune on tgt domain with src data}} \\ 
\midrule 
\texttt{BERT  fine-tune} & 64.97 & 95.15 & 85.09 & - & - & - & -\\ 
\texttt{Proto} & \textbf{77.02} & 95.29 & 89.20 & - & - & - & 86.31\\
\texttt{Retriever} & 56.36 & \textbf{97.17} & 83.56 & 76.87 & 94.11 & 88.36 &  \textbf{88.18}\\
\texttt{Retriever  min-max} & 58.82 & 96.85 & 84.17 & 74.31 & 94.32 & 87.56 & 83.98\\

\bottomrule
\end{tabular}
\end{center}
\caption{\label{clinc_results_10_10_2} Intent accuracy on CLINC for $n_c=10, n_i=10$ with 2-shots.}
\vspace{-1em}
\end{table*}

\begin{table*}[t]
\footnotesize
\begin{center}
\begin{tabular}{l|ccc|ccc|c}
\toprule
& \multicolumn{3}{c|}{support\_set=all} & \multicolumn{3}{c|}{support\_set=balance} & support\_set=tgt \\
\cmidrule(lr){2-4} \cmidrule(lr){5-7} \cmidrule(lr){8-8}
& tgt   & src  & avg   & tgt & src & avg   & tgt  \\ 
\midrule
\multicolumn{4}{l}{\textit{Initial BERT }}     \\
\midrule
$\texttt{Proto}^{\texttt{frz}}$  & 12.57 & 21.67 & 17.42 & - & - & - & -\\
$\texttt{Retriever}^{\texttt{frz}}$ & 9.46 & 56.96 & 34.79 & 23.37 & 25.87 & 24.70 & 32.72 \\
\midrule
\multicolumn{4}{l}{\textit{Pre-train on src domain }} \\  
\midrule
\texttt{BERT fine-tune} &- & 96.92 & - & - & - & - & -\\ 
\texttt{Proto} & 73.17 & 94.96 & 84.79 & - & - & - & -\\ 
\texttt{Retriever} & 66.56 & 96.76 & 82.67 & 76.56 & 94.99 & 86.39 & 83.72 \\
\texttt{Retriever  min-max} & 66.08 & 96.64 & 82.38 & 73.14 & 95.13 & 84.87 & 79.52 \\
\midrule
\multicolumn{4}{l}{\textit{Fine-tune on tgt domain }} \\
\midrule 
\texttt{BERT fine-tune} & 76.92 & 44.17 & 59.45 & - & - & - & -\\
\texttt{Proto} & 77.53 & 94.88 & 86.78 & - & - & - & 85.21\\
\texttt{Retriever} & 68.35 & \textbf{97.11} & 83.69 & 77.54 & 95.67 & 87.21 & 85.72\\
\texttt{Retriever  min-max} & 70.65 & 96.76 & 84.58 & 76.74 & 94.90 & 86.43 & 85.41\\
\midrule 
\multicolumn{4}{l}{\textit{Fine-tune on tgt domain with src data}} \\ 
\midrule 
\texttt{BERT  fine-tune} & 70.48 & 95.11 & 83.61 & - & - & - & -\\ 
\texttt{Proto} & 79.89 & 94.32 & 87.59 & - & - & - & 85.41 \\
\texttt{Retriever} & 71.46 & 96.88 & 85.02 & \textbf{79.90} & 94.61 & \textbf{87.75} & \textbf{89.67}\\
\texttt{Retriever  min-max} & 71.73 & 96.86 & 85.13 & 78.43 & 95.31 & 87.43 & 85.32\\

\bottomrule
\end{tabular}
\end{center}
\caption{\label{clinc_results_8_10_5} Intent accuracy on CLINC for $n_c=8, n_i=10$ with 5-shots.}
\vspace{-1em}
\end{table*}

\begin{table*}[t]
\footnotesize
\begin{center}
\begin{tabular}{l|ccc|ccc|c}
\toprule
& \multicolumn{3}{c|}{support\_set=all} & \multicolumn{3}{c|}{support\_set=balance} & support\_set=tgt \\
\cmidrule(lr){2-4} \cmidrule(lr){5-7} \cmidrule(lr){8-8}
& tgt   & src  & avg   & tgt & src & avg   & tgt  \\ 
\midrule
\multicolumn{4}{l}{\textit{Initial BERT }}     \\
\midrule
$\texttt{Proto}^{\texttt{frz}}$  & 7.24 & 20.99 & 14.57 & - & - & - & -\\
$\texttt{Retriever}^{\texttt{frz}}$ & 4.76 & 57.24 & 32.75 & 17.64 &	18.29 & 17.99 & 24.17 \\
\midrule
\multicolumn{4}{l}{\textit{Pre-train on src domain }} \\  
\midrule
\texttt{BERT fine-tune} & - & 96.92 & - & - & - & - & -\\ 
\texttt{Proto} & 70.59 & 94.42 & 83.30 & - & - & - & -\\ 
\texttt{Retriever} & 54.46 & \textbf{97.22} & 77.27 & 68.06 & 93.90 & 81.84 & 78.89\\
\texttt{Retriever  min-max} & 56.36 & 96.79 & 77.92 & 65.43 & 94.46 & 80.91 & 74.19\\
\midrule
\multicolumn{4}{l}{\textit{Fine-tune on tgt domain }} \\
\midrule 
\texttt{BERT fine-tune} & 68.94 & 56.41 & 62.25 & - & - & - & -\\
\texttt{Proto} & 73.19 & 94.70 & 84.66 & - & - & - & 79.83 \\
\texttt{Retriever} & 57.58 & 97.01 & 78.61 & 70.98 & 93.56 & 83.02 & \textbf{80.35} \\
\texttt{Retriever  min-max} & 58.71 & 96.89 & 79.07 & 68.71 & 94.64 & 82.54 & 76.51\\
\midrule 
\multicolumn{4}{l}{\textit{Fine-tune on tgt domain with src data}} \\ 
\midrule 
\texttt{BERT  fine-tune} & 61.53 & 95.10 & 79.43 & - & - & - & -\\ 
\texttt{Proto} & \textbf{73.57} & 94.59 & \textbf{84.78} & - & - & - & 80.08\\
\texttt{Retriever} & 59.06 & 97.07 & 79.33 & 72.33 & 93.11 & 83.42 & 80.14\\
\texttt{Retriever  min-max} & 60.49 & 96.89 & 79.91 & 70.92 & 94.68 & 83.59 & 79.21 \\

\bottomrule
\end{tabular}
\end{center}
\caption{\label{clinc_results_8_10_2} Intent accuracy on CLINC for $n_c=8, n_i=10$ with 2-shots.}
\vspace{-1em}
\end{table*}

\begin{table*}[t]
\footnotesize
\begin{center}
\begin{tabular}{l|ccc|ccc|c}
\toprule
& \multicolumn{3}{c|}{support\_set=all} & \multicolumn{3}{c|}{support\_set=balance} & support\_set=tgt \\
\cmidrule(lr){2-4} \cmidrule(lr){5-7} \cmidrule(lr){8-8}
& tgt   & src  & avg   & tgt & src & avg   & tgt  \\ 
\midrule
\multicolumn{4}{l}{\textit{Initial BERT }}     \\
\midrule
$\texttt{Proto}^{\texttt{frz}}$  & 15.02 & 17.58 & 16.30 & - & - & - & -\\
$\texttt{Retriever}^{\texttt{frz}}$ & 10.65 & 55.97 & 33.31 & 21.16 & 25.85 & 23.51 & 26.71\\
\midrule
\multicolumn{4}{l}{\textit{Pre-train on src domain }} \\  
\midrule
\texttt{BERT fine-tune} & - & 96.87 & - & - & - & - & -\\ 
\texttt{Proto} & 67.05 & 96.37 & 81.71 & - & - & - & -\\ 
\texttt{Retriever} & 64.19 & \textbf{97.18} & 80.69 & 68.40 & 95.85 & 82.13 & 72.02\\
\texttt{Retriever  min-max} & 61.79 & 96.71 & 79.25 & 65.11 & 95.85 & 80.48 & 68.21
\\
\midrule
\multicolumn{4}{l}{\textit{Fine-tune on tgt domain }} \\
\midrule 
\texttt{BERT fine-tune} & 69.32 & 64.76 & 67.04 & - & - & - & -
\\
\texttt{Proto} & 73.63 & 96.09 & 84.86 & - & - & - & 77.45
\\
\texttt{Retriever}& 68.65 & 97.05 & 82.85 & 74.59 & 96.24 & 85.42 & 77.96
\\
\texttt{Retriever  min-max} & 67.94 & 97.04 & 82.49 & 71.88 & 96.27 & 84.07 & 75.93
\\
\midrule 
\multicolumn{4}{l}{\textit{Fine-tune on tgt domain with src data}} \\ 
\midrule 
\texttt{BERT  fine-tune}& 67.93 & 95.95 & 81.94 & - & - & - & -
\\ 
\texttt{Proto}& 74.43 & 95.63 & 85.03 & - & - & - & 76.61
 \\
\texttt{Retriever}& 71.51 & 96.91 & 84.21 & \textbf{76.37} & 95.81 & \textbf{86.09} & \textbf{78.62}
\\
\texttt{Retriever  min-max} & 70.50 & 96.92 & 83.71 & 73.76 & 95.79 & 84.78 & 76.65
 \\

\bottomrule
\end{tabular}
\end{center}
\caption{\label{clinc_results_5_15_5} Intent accuracy on CLINC for $n_c=5, n_i=15$ with 5-shots.}
\vspace{-1em}
\end{table*}

\begin{table*}[t]
\footnotesize
\begin{center}
\begin{tabular}{l|ccc|ccc|c}
\toprule
& \multicolumn{3}{c|}{support\_set=all} & \multicolumn{3}{c|}{support\_set=balance} & support\_set=tgt \\
\cmidrule(lr){2-4} \cmidrule(lr){5-7} \cmidrule(lr){8-8}
& tgt   & src  & avg   & tgt & src & avg   & tgt  \\ 
\midrule
\multicolumn{4}{l}{\textit{Initial BERT }}     \\
\midrule
$\texttt{Proto}^{\texttt{frz}}$  & 9.39 & 16.19 & 12.79 & - & - & - & -\\
$\texttt{Retriever}^{\texttt{frz}}$ & 5.18 & 56.25 & 30.72 & 14.99 &	18.36 & 16.68 & 19.63 \\
\midrule
\multicolumn{4}{l}{\textit{Pre-train on src domain }} \\  
\midrule
\texttt{BERT fine-tune} & - & 96.87 & - & - & - & - & -
\\ 
\texttt{Proto} & 63.50 & 95.72 & 79.61 & - & - & - & -
\\ 
\texttt{Retriever} & 56.19 & \textbf{97.14} & 76.67 & 63.75 & 95.36 & 79.55 & 67.82
\\
\texttt{Retriever  min-max} & 55.88 & 96.94 & 76.41 & 60.32 & 95.50 & 77.91 & 63.60
\\
\midrule
\multicolumn{4}{l}{\textit{Fine-tune on tgt domain }} \\
\midrule 
\texttt{BERT fine-tune} & 61.91 & 77.11 & 69.51 & - & - & - & - \\
\texttt{Proto} & 66.38 & 95.72 & 81.05 & - & - & - & 70.40
\\
\texttt{Retriever} & 58.83 & 97.01 & 77.92 & 66.90 & 95.30 & 81.10 & \textbf{71.26}
\\
\texttt{Retriever  min-max} & 57.97 & 97.07 & 77.52 & 63.23 & 95.78 & 79.50 & 66.44
 \\
\midrule 
\multicolumn{4}{l}{\textit{Fine-tune on tgt domain with src data}} \\ 
\midrule 
\texttt{BERT  fine-tune} & 60.32 & 96.16 & 78.24 & - & - & - & -
\\ 
\texttt{Proto} & 66.98 & 95.77 & \textbf{81.38} & - & - & - & 70.28
\\
\texttt{Retriever} & 59.88 & 96.96 & 78.42 & \textbf{67.23} & 94.85 & 81.04 & 70.52
\\
\texttt{Retriever  min-max} & 60.08 & 97.08 & 78.58 & 65.45 & 95.73 & 80.59 & 68.06
\\

\bottomrule
\end{tabular}
\end{center}
\caption{\label{clinc_results_5_15_2} Intent accuracy on CLINC for $n_c=5, n_i=15$ with 2-shots.}
\vspace{-1em}
\end{table*}

\end{document}